\pdfoutput=1

\documentclass[11pt]{article}

\usepackage[preprint]{acl}

\usepackage{times}
\usepackage{latexsym}

\usepackage[T1]{fontenc}
\usepackage{algorithm}
\usepackage{algorithmic}
\usepackage[utf8]{inputenc}

\usepackage{microtype}

\usepackage{inconsolata}

\usepackage{graphicx}
\usepackage{booktabs}
\usepackage{url}
\usepackage{amsmath}
\usepackage{tcolorbox}
\usepackage{subfig}
\usepackage{comment}
\usepackage{makecell}
\title{Diagnosing Moral Reasoning Acquisition in Language Models:\\Pragmatics and Generalization}

\author{
  \textbf{Guangliang Liu\textsuperscript{1}}
~\textbf{Zimo Qi\textsuperscript{2}}
~\textbf{Xitong Zhang\textsuperscript{1}}\\
~\textbf{Lei Jiang\textsuperscript{3}}
~\textbf{Kristen Marie Johnson\textsuperscript{1}}
\\
  \textsuperscript{1}Michigan State University
~\textsuperscript{2}Johns Hopkins University\\
~\textsuperscript{3}University of Illinois Chicago
\\
\texttt{\{liuguan5,zhangxit,kristenj\}@msu.edu}\\~~~~\texttt{zqi15@jh.edu}~~~~\texttt{ljian43@uic.edu}
}

\begin{document}
\maketitle
\begin{abstract}
Ensuring that Large Language Models (LLMs) return just responses which adhere to societal values is crucial for their broader application. 
Prior research has shown that LLMs often fail to perform satisfactorily on tasks requiring moral cognizance, such as ethics-based judgments. While current approaches have focused on fine-tuning LLMs with curated datasets to improve their capabilities on such tasks, choosing the optimal learning paradigm to enhance the ethical responses of LLMs remains an open research debate. In this work, we aim to address this fundamental question:~\textit{can current learning paradigms enable LLMs to acquire sufficient moral reasoning capabilities?} Drawing from distributional semantics theory and the pragmatic nature of moral discourse, our analysis indicates that 
performance improvements follow a mechanism similar to that of semantic-level tasks, and therefore remain affected by the pragmatic nature of morals latent in discourse, a phenomenon we name the \textit{pragmatic dilemma}. We conclude that this pragmatic dilemma imposes significant limitations on the generalization ability of current learning paradigms, making it the primary bottleneck for moral reasoning acquisition in LLMs. 

\textit{\small \textbf{Warning}: examples in this paper may be offensive.}



\end{abstract}
\section{Introduction}
Given the widespread usage of LLMs across all facets of society, enabling such models with moral reasoning capabilities has become a significant research goal. Though AI alignment~\cite{bai2022training} has become the de-facto method to align LLMs with human values, its effectiveness has been debated~\cite{lin2023unlocking,qi2024safety}.
One significant complaint is that alignment with human preference does not allow LLMs to achieve intrinsic alignment, resulting in various safety issues, e.g., jailbreak attacks~\cite{xie2023defending} and propagation of social biases to downstream tasks~\cite{liu-etal-2024-intrinsic}.
However, enabling LLMs to develop moral reasoning capabilities is a non-trivial task; it is both a pragmatics-level task~\cite{awad2022computational}, as well as philosophically challenging, due to debate over the correct representation of human morals and ethics~\cite{zhixuan2024preferencesaialignment}.

\citet{jiang2021can} and~\citet{hendrycks2020aligning} represent pioneering efforts to enable LLMs to acquire ethical judgment capabilities by fine-tuning them on curated textual data that jointly depicts various moral situations alongside corresponding judgments.
\citet{zhou2024rethinking} introduces an in-context learning method to help LLMs perform moral reasoning, based on a top-down framework driven by the Moral Foundation Theory~\cite{anderson2007machine}.
~\citet{liutraining} introduce a social sandbox wherein LLMs can learn how to be moral through interactions.
~\citet{tennant2024moral} propose a moral alignment framework to make LLM agents behave morally through a newly designed intrinsic moral reward function based on the Iterated Prisoner's Dilemma\footnote{\url{https://en.wikipedia.org/wiki/Prisoner\%27s_dilemma}}.
In addition to those efforts proposing solutions, new benchmarks have also been proposed~\cite{forbes2020social,hendrycks2020aligning,ren-etal-2024-valuebench}.

There are also several studies which highlight the inefficiency of LLMs on tasks requiring moral reasoning. 
\citet{talat2022machine} has criticized the \citet{jiang2021can} work described above, because while their intended goal was normative ethics, they instead leveraged a bottom-up approach for learning descriptive ethics~\cite{vida-etal-2023-values,fraser2022does}.
\citet{jin2022make} empirically demonstrate that the current learning paradigm for moral reasoning tasks relies on a large training dataset. \citet{sap2022neural} also show the failure of LLMs on social intelligence tasks such as theory-of-mind.

In cognitive science, \citet{mahowald2024dissociating} suggest that while LLMs excel in formal language competence, they struggle with functional language competence which is an essential requirement for acquiring moral reasoning capabilities.
More fundamentally,~\citet{bender2020climbing} and other studies in BERTology~\cite{rogers2021primer}, argue that Transformers cannot achieve true language acquisition, as it necessitates physical grounding and situated communicative intent~\cite{beuls2024humans}, which extends beyond the distributional semantics captured by Transformers~\cite{harris1954distributional,lenci2008distributional,boleda2020distributional}. 
Previous studies~\cite{bonagiri2024sage,zhang2023measuring} demonstrate that LLMs do not have consistent moral or ethical orientations across various instances, which is contrary to the moral consistency principle~\cite{arvanitis2020consistency}.
Appendix~\ref{sec:relatedworks} contains additional related works and motivation pertaining to machine ethics.

To address this debate, in this paper we pursue a deeper understanding of the mechanisms underlying current learning paradigms for moral reasoning acquisition. We argue that while existing paradigms can improve LLMs’ performance on morality-related tasks, this enhancement: (1) primarily arises from distributional similarities between seen and unseen ethical situations, and (2) faces challenges in generalization due to the inherently pragmatic nature of morality.
We name this phenomenon as the \textit{pragmatic dilemma}~\cite{laverick2010ethical,sap2022neural} of moral reasoning acquisition, which arises from the inherent gap between the nature of distributional semantics in LLMs and the pragmatic nature of morality. Significant consequences of this pragmatic dilemma include poor generalization and a lack of intrinsic alignment.

Specifically, we employ three fundamental tasks, Moral Foundations classification, rule of thumb generation, and ethical judgment prediction, as downstream evaluations of moral reasoning acquisition. We then compare their generalization characteristics with a representative semantics-driven task, sentiment analysis.
Motivated by the distributional semantics theory, we:
(1) empirically show the generalization and convergence pitfalls of Moral Foundations classification; (2) given the characteristic of autoregressive language models, propose a Representational Likelihood Algorithm (RLA) to statistically correlate \textit{representational similarity} between seen and unseen situations with the \textit{prediction likelihood} of unseen situations; and
(3) using RLA, perform mechanistic analysis of LLM performance gains for unseen situations.

Section~\ref{sec:preliminary} introduces the prevalent learning paradigm for moral reasoning acquisition and highlights the generalization challenges in fine-tuning masked language models for moral foundation prediction. Section~\ref{sec:moraltuning} presents experimental results across different learning paradigms, and Section~\ref{sec:mechanism} provides a detailed mechanistic analysis. Based on our experimental results, we conclude that the pragmatic dilemma blocks the effectiveness of current learning paradigms.
\section{Preliminary Background\label{sec:preliminary}}
In this section, we begin by introducing the benchmarks and dataset annotation used in our study. We then present the prevailing learning paradigm for moral reasoning acquisition. Finally, we use the Moral Foundations prediction task with a Masked Language Model, specifically BERT~\cite{devlin2019bert}, as a case study, to illustrate the generalization challenges of this task by drawing comparisons to the semantics-level task of sentiment analysis.
\subsection{Benchmark and Dataset Annotation\label{subsec:dataset}}
\begin{table}[h]
    \centering
    \small
    \begin{tabular}{l}
        \toprule
        \textbf{Situation}: Reminding my coworker who crashed \\into my car to pay to get it repaired. \\
        \midrule
        \textbf{Moral Foundation}: Fairness.\\
        \midrule
        \textbf{Rule of Thumb (RoT)}: If you crash into someone's car,\\ you should pay for their repairs.\\
        \midrule
        \textbf{(Ethical) Judgment}: You should.\\
        \bottomrule
    \end{tabular}
    \caption{\small Dataset Annotation. Given a moral situation describing a morality-relevant case, the corresponding Moral Foundation, RoT, and Judgment are presented.}
    \label{tab:dataset}
\end{table}
We employ two popular benchmarks: MIC~\cite{ziems2022moral} and SocialChem~\cite{forbes2020social}.
Table~\ref{tab:dataset} presents an overview of the dataset annotations used across both benchmarks.
Given a moral situation, the \textit{Moral Foundation}~\cite{haidt2004,haidt2007morality} represents the underlying social norm that the situation either adheres to or violates (please refer to Table~\ref{appendix:MFs} for more details of Moral Foundation Theory). The \textit{RoT (Rule of Thumb)} encapsulates a fundamental explanation of right and wrong behavior, serving as a guidance for the subsequent ethical judgment. The \textit{(Ethical) Judgment} then determines whether the given situation is deemed acceptable or unacceptable.
While a single moral situation may be associated with multiple moral foundations, this study focuses exclusively on cases where only one underlying moral foundation is present.
In the MIC, each prompt-reply pair is treated as a distinct situation.

\subsection{Learning Paradigms\label{subsec:learning-paradigm}}
Existing learning paradigms for moral reasoning acquisition generally fine-tune LLMs on curated textual data that depicts various moral situations alongside corresponding judgments or actions. 
In previous studies, ethical judgment prediction and RoT generation are the most popular tasks~\cite{bonagiri2024sage,ren-etal-2024-valuebench,hendrycks2020aligning,sorensen2024value}, and Moral Foundations classification is widely accepted in the area of computational social science~\cite{johnson2018classification,roy2021identifying}. 
Though there is no agreed-upon definition for moral reasoning acquisition, we consider Moral Foundations classification, RoT generation, and ethical judgment prediction as three downstream tasks indicative of moral reasoning capabilities.
Although some studies incorporate interactive sandboxes or multi-round feedback into learning paradigms~\cite{liutraining,wang-etal-2024-sotopia}, Moral Foundations classification, RoT generation, and ethical judgment prediction remain fundamental tasks, which when fine-tuned with LLMs form the preferred learning paradigms. 

\textbf{Notations.} We denote the moral situation as $m_s$, the moral foundation as $y_m$, the RoT as $y_r$, and the judgment as $y_j$.
Assuming an LLM $f$ is parameterized by $\theta$, RoT generation is formulated as $y_r=f_{\theta}(m_s)$ and judgment prediction is represented as $y_j=f_{\theta}(m_s)$.

\textbf{Fine-tuning Strategies.} Current learning paradigms of moral reasoning acquisition which aim to maximize conditional probabilities $\mathcal{P}_{\theta}(y_r|m_s)$ and $\mathcal{P}_{\theta}(y_j|m_s)$, typically apply a self-supervised fine-tuning or a reinforcement learning loss objective\footnote{Please note the choice of objective loss function does not impact our conclusion.}.
Given the causal relationships among moral foundations, RoT, and judgment, previous studies often integrate them into a unified prediction task, such as $y_r=f_{\theta}(m_s, y_m)$ and $y_j=f_{\theta}(m_s,y_m,y_r)$.
During fine-tuning, the input for RoT generation can be $m_s$ with or without $y_m$, while the input for ethical judgment prediction can be $m_s$ with or without $y_m$ and/or $y_r$.

\subsection{Pitfalls of Generalization\label{subsec:pitfall-generalization}}
In this section, we use the Moral Foundations classification task as an example to illustrate its generalization pitfalls by comparing it to the semantics-level task of sentiment analysis.
We argue that \textit{in the moral foundations classification task, there should be serious generalization issues since the classification model has to map semantically different situations into the same moral foundation label}. 
A direct consequence is that an unseen situation is likely to be predicted correctly only if a semantically similar sample exists in the training set. This similarity requirement is much stricter than that for the sentiment analysis task.
\begin{table}[h]
    \centering
    \small
    \begin{tabular}{l}
        \toprule
        \textbf{Situation}: Kicking a kid out of his birthday party. \\
        \midrule
        \textbf{Situation}: Not telling my mom I smoke weed.\\
        \bottomrule
    \end{tabular}
    \caption{\small Situation Examples. Two moral situations with the same underlying moral foundation of authority-subversion.}
    \label{tab:situation2moral}
\end{table}

Our argument is driven by the gap between the distributional semantics captured by neural language models and the inherently pragmatic nature of morality.
For instance, Table~\ref{tab:situation2moral} presents two moral situations from the SocialChem benchmark; they are semantically different (\textit{distributional semantics}) but the underlying moral foundations are identical (\textit{pragmatics}).
If we force an MLM to map these two situations into the same moral foundation label, it would violate the captured distributional semantics during pre-training.  
To illustrate how the violation works, we refer to a semantics-level task of sentiment analysis using the SST dataset from the GLUE benchmark~\cite{wang2018glue}.

\textbf{Experimental Settings for Classification.} 
We have two settings for the moral classification tasks: classify moral situations to moral foundations and classify RoTs to moral foundations.
We use a fine-tuning dataset with 7500 randomly sampled cases and the bert-base-uncased\footnote{\url{https://huggingface.co/google-bert/bert-base-uncased}} model as the backbone model.
Beyond the backbone model, we insert a fully-connected layer as the classifier layer. More details about the hyperparameters setting is available in Appendix~\ref{appendix:hyperparams4clf}.
\begin{figure*}[t]
    \centering
    \includegraphics[width=0.98\linewidth]{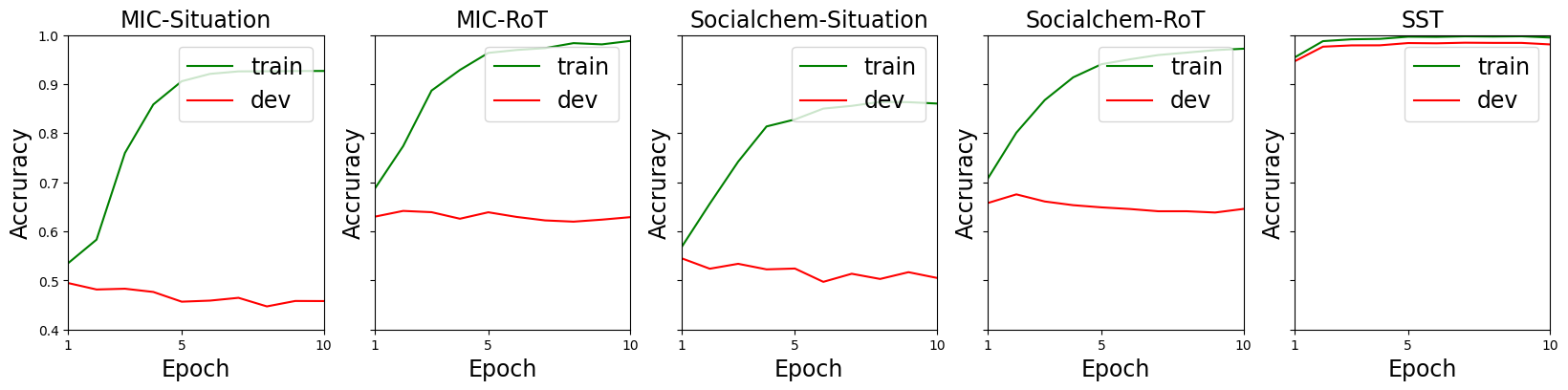}
    \caption{\small Training and Development Accuracy Over 10 Fine-tuning Epochs. The first four figures display results for moral foundation classification tasks, while the rightmost figure shows the results for the SST benchmark.}
    \label{fig:clf}
\end{figure*}

\textbf{Observations for Classification Performance.} Figure~\ref{fig:clf} presents the classification performance on both the training and development set. 
Compared to the generalization behavior observed in SST (rightmost figure), the moral foundation classification tasks (first four figures) exhibit a significant performance gap between the training set and the development set. 
However, for \texttt{MIC-RoT} and \texttt{SocialChem-RoT}, because the training accuracy approaches 100\% and converges after only several epochs, this suggests that \textit{task difficulty is not the primary cause of the observed generalization gap}.
The difference in classification performance between Situation and RoT stems from the fact that RoT is constructed based on typical moral foundations, inherently conveying information about the corresponding moral foundation. However, the generalization gap between the training set and development set for all moral foundation classification settings is apparent.
To further analyze the generalization pitfall in moral foundation classification, we examine the convergence behavior with respect to training dataset size. 
We use the curve of development accuracy in SST as a reference to highlight the convergence issue observed in moral foundation classification tasks.
\begin{figure}[h]
    \centering
    \includegraphics[width=0.8\linewidth]{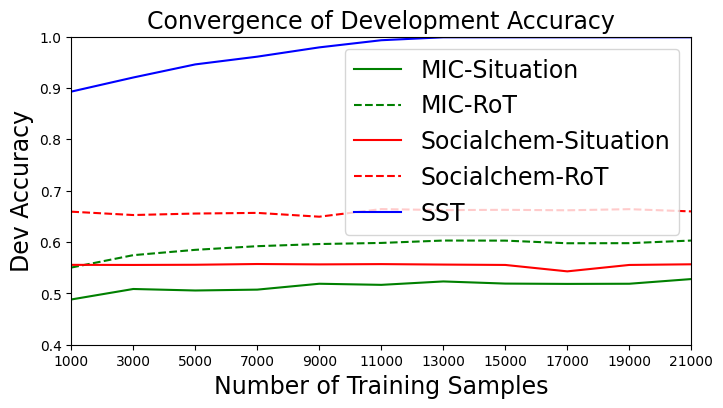}
    \caption{\small Convergence Curve of Development Accuracy for Considered Classification Tasks. Only the development accuracy of SST increases with more training samples and finally approaches 1.0.}
    \label{fig:convergence}
\end{figure}

\textbf{Experimental Settings for Convergence.} 
SST is a binary classification task. To ensure a fair comparison, we re-categorize the moral foundation labels for MIC and SocialChem to convert them into a binary classification task (details are in Appendix~\ref{appendix:redist_moral}).
For each task setting, we incrementally increase the training set size from 1,000 to 210,000 in steps of 2,000 and report the best performance on the development set at each training size setting.

\textbf{Observations for Convergence.} Figure~\ref{fig:convergence} illustrates the curve of development accuracy across all evaluated classification tasks. 
For SST, accuracy improves as the number of training samples increases, eventually stabilizing and approaching 1.0.
In contrast, the development accuracies for moral foundation classification tasks show no improvement in SocialChem and only marginal gains in MIC.
We believe this disparity is due to the fact that moral situations in SocialChem are generally shorter than that of MIC.
The convergence behavior analysis again showcases the generalization pitfalls of the moral foundation classification task.

In summary, we: (1) introduce the current learning paradigms for moral reasoning acquisition; and (2) show the generalization pitfalls of the moral foundation classification task (a pragmatics-level task) by referring and comparing to the development accuracy of a semantics-level task.

\section{Fine-tuning for Moral Reasoning Acquisition\label{sec:moraltuning}}
In this section, we introduce fine-tuning strategies and experimental results of existing learning paradigms for moral reasoning acquisition, focusing on the tasks of RoT generation and ethical judgment prediction.
\begin{table*}[t]
\small
\centering
\begin{tabular}{c c c c c| c c c c c}
\toprule
\texttt{SocialChem} &\text{\small BertScore} & \text{\small Rouge1} & \text{\small Rouge2} & \text{\small RougeL} & \texttt{MIC} &\text{\small BertScore} & \text{\small Rouge1} & \text{\small Rouge2} & \text{\small RougeL} \\ 
        \midrule
        rot&.777 & .229 & .096 & .213& rot&.768 & .175 & .077 & .168 \\ 
        \textbf{moral-rot}& .836 & .416 & .205 & .401& \textbf{moral-rot}&.826 & .393 & .192 & .379 \\ 
        \midrule
        judg&.7240 & .230 & .137 & .230 & judg&.671 & .071 & .000 & .071 \\ 
        \textbf{moral-judg}& .7632 & .464 & .346 & .464& \textbf{moral-judg}&.762 & .314 & .000 & .314 \\
        rot-judg&.7626 & .464 & .346 & .464& rot-judg&.660 & .061 & .000 & .061 \\
        \makecell{moral-rot\\-judg}&.7628 & .463 & .345 & .463& \makecell{moral-rot\\-judg}&.761 & .306 & .000 & .306 \\
        \bottomrule
\end{tabular}
\caption{\small Performance of Fine-tuned \texttt{Mistral} Model Across Various Fine-tuning Strategies for Each Benchmark, with the best strategy highlighted in \textbf{bold}. For both tasks, introducing more information, e.g., moral foundation, in the fine-tuning process would improve the performance. The~\texttt{moral-rot} achieves the optimal performance for both SocialChem and MIC. The~\texttt{moral-judg} is the best strategy for both SocialChem and MIC, in terms of the judgment prediction task. Additional results for Llama3 are availabe in Table~\ref{tab:ethicaltuning4llama3}.}
\label{tab:ethicaltuning4mistral}
\end{table*}

\textbf{Experimental Settings.}
We take Mistral-7B\footnote{\url{https://huggingface.co/mistralai/Mistral-7B-v0.1}} and Llama3-8B\footnote{\url{https://huggingface.co/meta-llama/Meta-Llama-3-8B-Instruct}} as the backbone models and leverage the LoRA method to fine-tune them through a supervised fine-tuning loss.
For each benchmark, we employ two fine-tuning strategies for RoT generation and four fine-tuning strategies for ethical judgment prediction. For \textit{RoT generation}, we fine-tune LLMs: (1) to directly generate RoT according to the given situation (\text{rot}) and (2) first generate the moral foundation, then the RoT (\text{moral-rot}). For \textit{Judgment Prediction}, we fine-tune LLMs to: (1) directly predict judgment (\text{judg}) , and (2) firstly generate the moral foundation and/or RoT then the judgment (\text{moral-judg},~\text{rot-judg} and~\text{moral-rot-judg}).

The prompting format and LoRA fine-tuning settings are available in Appendix~\ref{appendix:ethicaltuning}. 
We consider 10000 samples with only one underlying moral foundation for analytical convenience. In the process of fine-tuning, we take the check point with the least loss on the  development set, and report its performance on the test set.
During inference, we prompt fine-tuned LLMs to first generate intermediate predictions before producing the final RoT or ethical judgment, following the same prompting strategy used during fine-tuning.
For example, in the~\texttt{moral-rot} strategy, LLMs are instructed to first predict the moral foundation based on the given situation and subsequently generate the RoT using both the situation and the predicted moral foundation. Following~\citet{ziems2022moral}, we report the performance of the BertScore~\cite{zhang2019bertscore}, Rouge-1, Rouge-2, and Rouge-L metrics.

RoT generation and ethical judgment prediction align with the core capabilities essential for morality-related scenarios and serve as prototypical formats for moral reasoning.
By incorporating moral foundations into RoT generation, we aim to guide LLMs to first identify the moral foundation associated with a given situation, thereby improving the quality of the generated RoT. RoTs serve as instances of evidence and explanation for ethical judgments, aligning with previous studies that seek to enhance LLMs’ social intelligence through social interaction environments~\cite{liutraining,wang-etal-2024-sotopia}.

\textbf{Main Results.}
Table~\ref{tab:ethicaltuning4mistral} and Table~\ref{tab:ethicaltuning4llama3} present fine-tuning results for Mistral and Llama3, respectively\footnote{Note that this paper does not aim to achieve state-of-the-art performance but rather to investigate the underlying mechanisms behind these performance gains.}.
As shown in Table~\ref{tab:ethicaltuning4mistral}, introducing moral foundations in fine-tuning enhances performance across all experimental settings.
However, incorporating RoT information along into the ethical judgment prediction task has a negative impact to the MIC benchmark. We hypothesize that this is because judgments are significantly shorter than RoTs, and the added complexity of RoTs would introduce challenges for fine-tuning.
\section{Mechanistic Analysis\label{sec:mechanism}}

In the previous sections, we introduced preliminary studies regarding the generalization pitfalls of the moral foundations classification task (Section~\ref{sec:preliminary}), and the performance of fine-tuning LLMs for two moral reasoning tasks (Section~\ref{sec:moraltuning}).
In this section, we: 
(1) propose the Representational Likelihood Algorithm (RLA) which can uncover supportive training samples for a given test sample;
(2) explore the characteristics of supportive training samples, demonstrating that the introduction of additional information to enhance generalization aligns with the generalization mechanism of the semantics-level task;
(3) showcase that the \textit{pragmatic dilemma} still holds even though fine-tuned LLMs perform better in RoT generation and ethical judgment prediction.

\textbf{Motivation.}
Our study builds on the representational learning nature of LLMs and the widely accepted principle in generalization theory that a well-trained machine learning model can generalize effectively when the training and test set distributions are closely aligned in the feature space~\cite{zhou2022domain,hupkes2022state}.
Since neural language models capture distributional semantics, representational similarity can be interpreted as equivalent to distributional similarity.
Recall from Section~\ref{sec:preliminary} that we highlighted the generalization pitfalls of the moral foundation classification task. We argue that similar pitfalls should also exist for RoT generation and ethical judgment prediction.
Our hypothesis is that \textit{for a given test sample, the LLM can generalize effectively only if highly similar training samples have been adequately learned during fine-tuning}. To test this hypothesis, we propose a novel algorithm to identify the training samples most conducive to the generalization of a given test sample within the representation space.
\subsection{Representational Likelihood Algorithm\label{subsec:hypo}}
Motivated by the representation similarity hypothesis in domain generalization~\cite{ben2006analysis}, we present our method for identifying training samples that contribute to the prediction of a given test sample. We refer to these training samples as \textit{generalization-supportive samples}\footnote{In this paper, we use generalization-supportive and supportive interchangeably.}. Our goal is to correlate representational similarity with LLM predictions, and then leverage this correlation to characterize the generalization mechanism of the considered morality acquisition tasks.

Assume that a fine-tuned LLM $f_{\theta}$ has been trained on the training set $\mathcal{D}_{\text{train}}$, where each sample is represented as $x = [m_s,y_m,y_r,y_j]$,
following the annotation introduced in Section~\ref{subsec:learning-paradigm}.
We denote training samples as $x \sim \mathcal{D}_{\text{train}}$ and test samples as $x' \sim \mathcal{D}_{\text{test}}$.
The hidden states of $f_{\theta}$ are denoted by $\mathcal{H}_{\theta}(\cdot)$, and the conditional likelihood of a given input and output is represented as $\mathcal{P}_{\theta}(\cdot|\cdot)$.
Denote the cosine similarity function as $\text{cos}(\cdot)$.
\begin{algorithm}
\caption{RLA for Judgment Prediction}
\label{alg:simulation}
\begin{algorithmic}[1] 
    \STATE Initialize $r = 0$, $\mathbf{d}=\{\}$
    \FOR{each sample $x'$ in $\mathcal{D}_{\text{test}}$}
        \STATE Sampling $\mathcal{N}$ cases from $\mathcal{D}_{\text{train}}$ as\\ $\mathcal{X}=[x^1,x^2,\cdots,x^{\mathcal{N}}]$\\
        \FOR{each $x^t$ in $\mathcal{X}$} 
            \STATE $S^t$ = $ \overbrace{\text{cos}(\mathcal{H}_{\theta}(m_s^t),\mathcal{H}_{\theta}(m_s^{'}))}^{\text{\textbf{representational similarity}}}$ $\cdot$ $\overbrace{\mathcal{P}_{\theta}(y^t_{j}|m^{t}_{s})}^{\textbf{likelihood}}$
            \STATE $\mathbf{d}[S^t]=\underbrace{\mathcal{P}_{\theta}(y^t_{j}|m^{'}_{s})}_{\text{\textbf{prediction}}}$
        \ENDFOR
        \STATE Sort $\mathbf{d}$ by key in \textit{ascending} order, return the value list as $\mathcal{V}$
        \IF{\text{MEAN}($\mathcal{V}$$[:\frac{\mathcal{N}}{2}]$) $<$ \text{MEAN}($\mathcal{V}[\frac{\mathcal{N}}{2}:]$) }
            \STATE $r$++ 
       
        \ENDIF
    \ENDFOR
    \STATE \textbf{return} $\frac{r}{\#\mathcal{D}_{\text{test}}}$
\end{algorithmic}
\end{algorithm}
Algorithm~\ref{alg:simulation} presents our proposed Representational Likelihood Algorithm (RLA) by taking the \texttt{judg} fine-tuning strategy ($y_j=f_{\theta}(m_s)$) as an instance. 
Specifically,\begin{itemize}
\setlength{\itemsep}{0pt}
\setlength{\parsep}{0pt}
\setlength{\parskip}{0pt}
    \item[1.] For each test case, we randomly sample $\mathcal{N}$ samples $\mathcal{X}$ from the training set (line 3).
    \item[2.] For each training sample $x^t$ in the sampled set $\mathcal{X}$, we calculate the \textbf{similarity score $S^t$} which comprises the: (1) cosine similarity between two hidden states $\mathcal{H}_{\theta}(m_s^t)$ and $\mathcal{H}_{\theta}(m_s^{'})$ (line 5) measuring the representational similarity, and (2) likelihood, the conditional probability $\mathcal{P}_{\theta}(y^t_{j}|m^{t}_{s})$ measuring how good $f_{\theta}$ fits $x^{t}$ (line 5). With this design, only those training samples that have been fitted well by $f_{\theta}$ would be considered in the process of measuring representational similarity.
    \item[3.] Compute the conditional probability of the training sample’s judgment given the test case’s situation (line 6).
    \item[4.] If $f_{\theta}$ becomes increasingly likely to assign $m_s^t$'s judgment $y^t_j$ to $m'_s$ as their representational similarity increases, then we can correlate representational similarity and prediction (lines 8-10).
\end{itemize}

In our experiments, we utilize the hidden states from the ${15}^{th}$ layer onward of the final token as the representation and compute the average cosine similarity across these layers to obtain the representational similarity score. This is because previous studies~\cite{geva2023dissecting,liu-etal-2024-intrinsic} indicate that the LLMs considered in this paper generally exhibit differences in the hidden state space from the ${15}^{th}$ layer onward.
\begin{table}[ht]
\small
    \centering
    \begin{tabular}{c c c}
        \toprule
        &Mistral & Llama3 \\ 
        \midrule
        Socialchem-rot&.920 & .924 \\ 
        \midrule
        Socialchem-judg& .998 & .996 \\ 
        \midrule
       MIC-rot& .926 & .912 \\ 
        \midrule
        MIC-judg& .990 & .971 \\ 
        \bottomrule
    \end{tabular}
    \caption{\small Experimental results for the simulation task show that all values exceed 0.9, indicating a strong correlation between representational similarity and prediction.}
    \label{tab:distri_hypo}
\end{table}
Table~\ref{tab:distri_hypo} presents the results of two baseline fine-tuning strategies, \texttt{rot} and \texttt{judg}, evaluated across various benchmarks and LLM models.
As shown, all experimental results exceed 0.9, particularly the \texttt{judg} fine-tuning strategy which is very close to 1.0, demonstrating that there exists correlation between representational similarity and prediction. 
In other words, for a given test sample, generalization-supportive training samples can be identified by assessing their representational similarity.
\subsection{Interpretation of Generalization}

Building on the method for identifying generalization-supportive training samples from Section~\ref{subsec:hypo}, this section interprets the generalization mechanism of the examined morality-relevant tasks by analyzing the characteristics of these supportive training samples\footnote{In this section, we provide a detailed analysis only for the fine-tuned Mistral, while the analysis for the fine-tuned Llama3 is presented in Appendix~\ref{appendix:mech_llama3}.}.

\begin{figure}[t]
    \centering
    \subfloat{\includegraphics[width=0.99\linewidth]{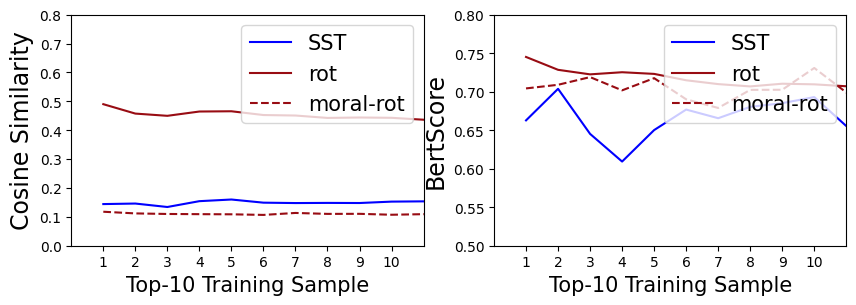}}
    \hfill
    \subfloat{\includegraphics[width=0.99\linewidth]{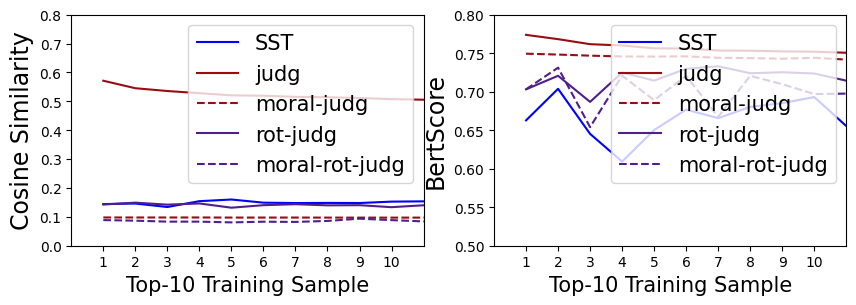}}
    \hfill
    \subfloat{\includegraphics[width=0.99\linewidth]{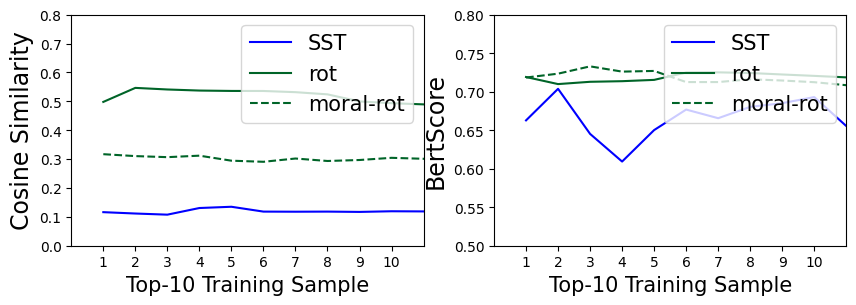}}
    \hfill
    \subfloat{\includegraphics[width=0.99\linewidth]{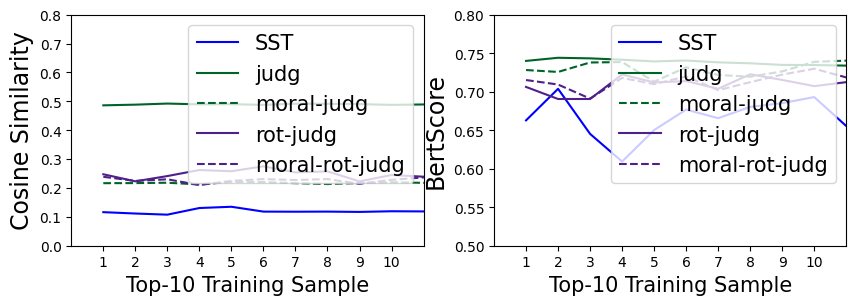}}
    \hfill
    \caption{\small Top-10 generalization-supportive training samples analysis for fine-tuned Mistral with the SocialChem (upper two rows) and MIC (bottom two rows) benchmark. 
    }
    \label{fig:mech_mistral}
\end{figure}

For each test sample, we collect the top-10 generalization-supportive training samples with the most highest similarity score $S^t$.
However, the similarity score $S^t$ is a high-level metric capturing the statistical correlation between representational similarity and predictions, making it insufficient for directly interpreting the underlying reasons for performance gains.
To have an in-depth analysis, we investigate (i) the cosine similarity of hidden states between the test sample's moral situation and the training sample's moral situation; (ii) the BertScore between the train sample's situation and the test sample's situation. 
Figure~\ref{fig:mech_mistral} present these two analytical perspectives on the top 10 generalization-supportive training samples for the fine-tuned Mistral model across two benchmarks.

By zooming into the left four subfigures in Figure~\ref{fig:mech_mistral}, introducing moral foundation or RoT in the fine-tuning process can decrease the representational similarity, particularly the optimal fine-tuning strategies, e.g., \texttt{moral-rot} and \texttt{moral-judg}, lead to lower representational similarities than that of the baseline strategy (\texttt{rot} and \texttt{judg}).
This phenomenon aligns with our hypothesis that \textit{generalization in moral reasoning acquisition tasks requires a high degree of representational similarity between test and training samples}.

By referring to the curve of SST that also faces a lower representational similarity, we can conclude that additional information of moral foundation or RoT would alleviate the generalization pitfall of the baseline strategy that necessitates much similar training samples to generalize.
This is rather natural since those fine-tuning strategies not only capture the information of situations but also moral foundations and/or RoTs, newly introduced information would impact the characteristics of the representation space.

Additionally, we can observe decreased BertScore in the right four sub-figures, except for RoT generation in the MIC benchmark, where the BertScore for \texttt{moral-rot} remains close to that of the baseline \texttt{rot} strategy.
A decrease in BertScore suggests that the additional information reduces reliance on generalization-supportive training samples with high distributional similarity to the test sample.
Due to the association between distributional similarity and representational similarity in LLMs, those two observations are aligned.
It is not surprising that the performance gain arises from the generalization mechanism analogical to that of semantics-level tasks. 
A natural question is~\textit{does the incorporation of moral foundations or RoT alleviate the pragmatic dilemma of current learning paradigms in moral reasoning acquisition?}

\begin{figure}[t]
    \centering
    \begin{minipage}{0.23\textwidth}
        \centering
        \includegraphics[width=\linewidth]{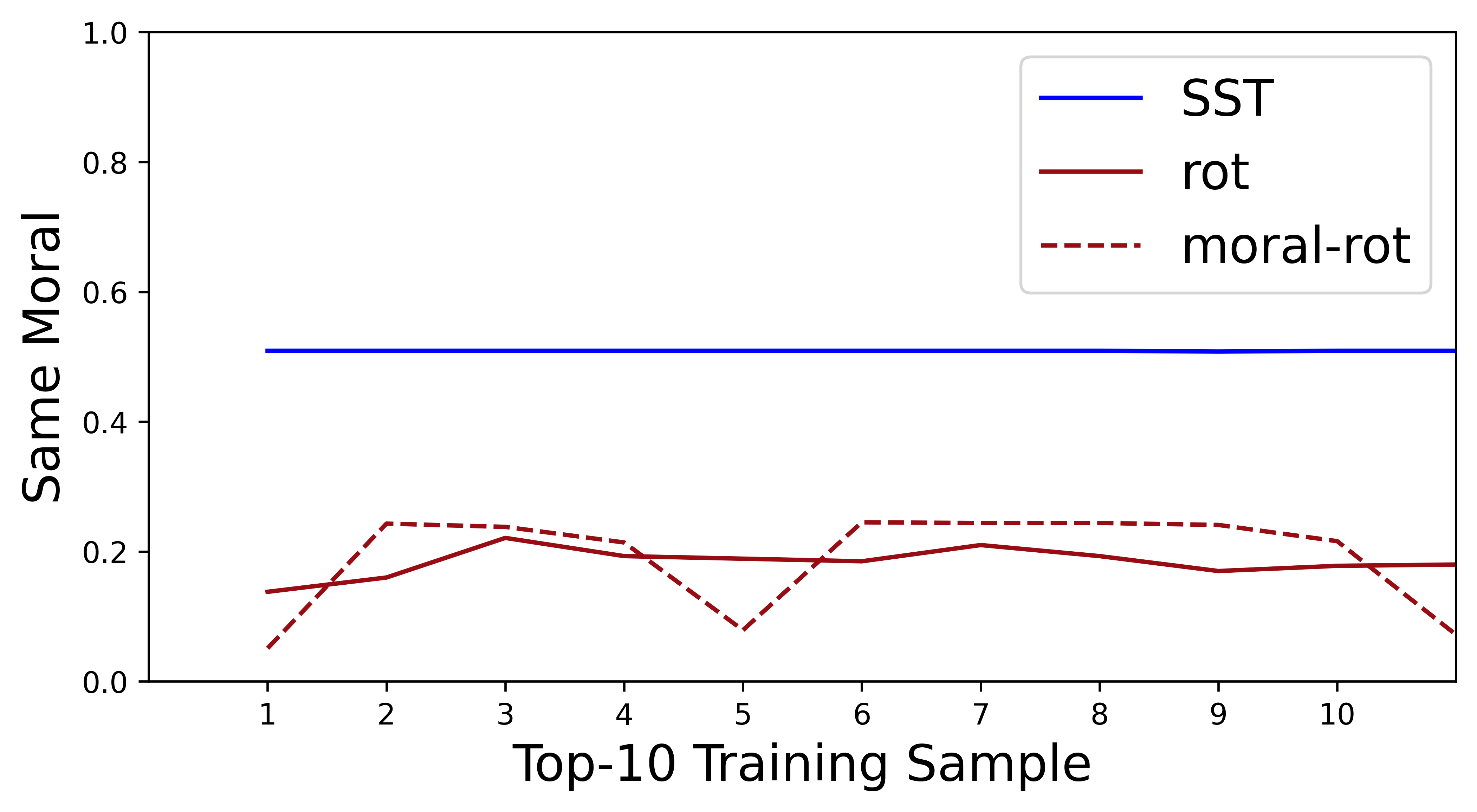}
    \end{minipage}
    \hfill
    \begin{minipage}{0.23\textwidth}
        \centering
        \includegraphics[width=\linewidth]{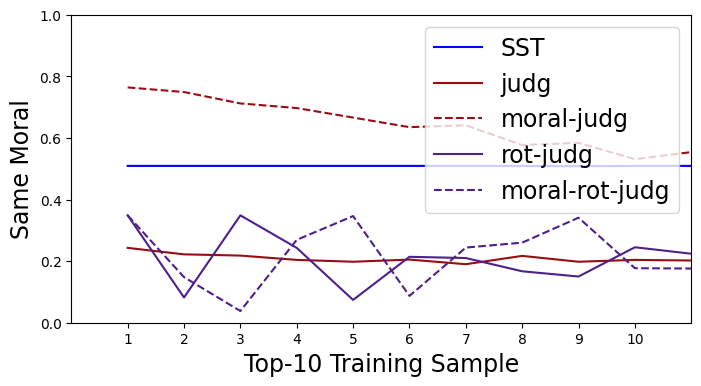}
    \end{minipage}
    \hfill
    \begin{minipage}{0.23\textwidth}
        \centering
        \includegraphics[width=\linewidth]{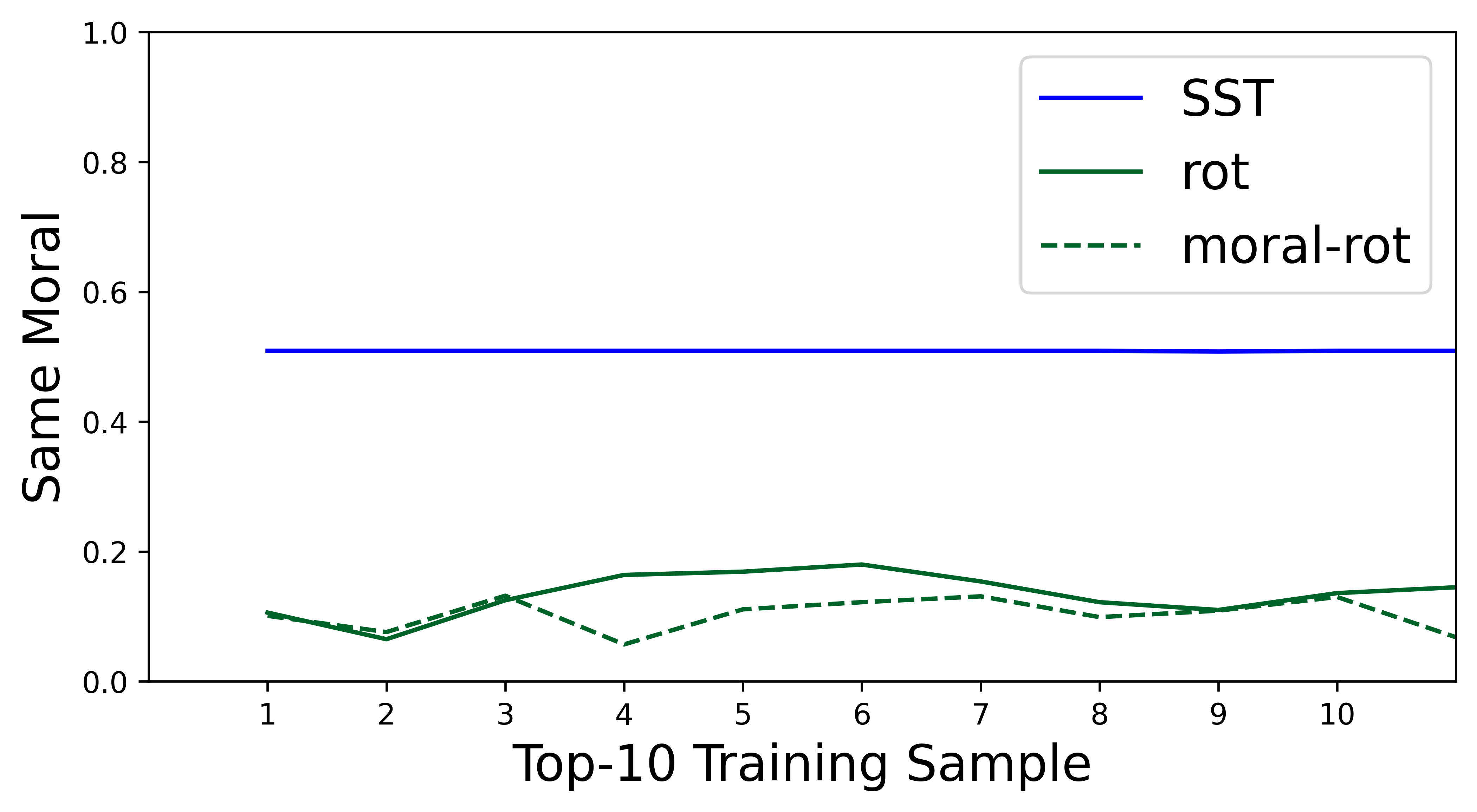}
    \end{minipage}
    \hfill
    \begin{minipage}{0.23\textwidth}
        \centering
        \includegraphics[width=\linewidth]{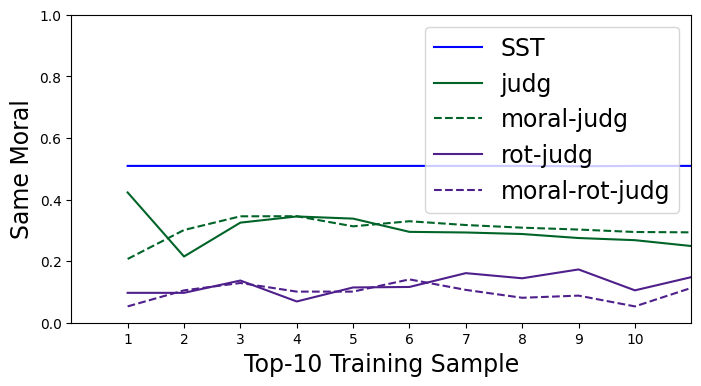}
    \end{minipage}
    \hfill
    \caption{Ratio of generalization-supportive training situations with the same underlying moral foudation as the test situation. Upper two subfigures are for SocialChem and the bottom sub-figures are for MIC. Top-50 situations are available in Appendix~\ref{app:top100}.}
    \label{fig:same_moral_mistral}
\end{figure}
An extreme case for the vanishment of the pragmatic dilemma is: \textit{for a given test situation, top-10 generalization-supportive training moral situations should have the same underlying moral foundations as the test moral situation}.
Therefore, we compute the ratio of the top-10 supportive training moral situations that share the same moral foundations as the test moral situation. 
Notably, we take the term training/test moral situation, for MIC and SocialChem, instead of training/test samples to emphasize that our analysis exactly focuses on moral situations.
For reference, we include SST and consider the sentiment label when calculating the ratio for SST.

Figure~\ref{fig:same_moral_mistral} presents the results for this ratio.
Interestingly, even for SST, which can be viewed as a binary classification task, only half of the supportive training samples share the same sentiment label as their corresponding test samples.
For both RoT generation and ethical judgment prediction, the optimal fine-tuning strategies (\texttt{moral-rot} and \texttt{moral-judg}) align with the baseline fine-tuning strategies (\texttt{rot} and \texttt{judg}), except for \texttt{moral-judg} on the SocialChem benchmark.
We believe this exception arises because the textual length of moral situations in SocialChem is relatively short, amplifying the influence of ethical judgment during fine-tuning.

On the other hand, Table~\ref{tab:likelihoods} reports the average conditional likelihoods of the top-10 supportive training situations, and note the optimal fine-tuning strategy does help LLMs fit training samples. These observations suggest that LLMs consider moral situations and additional information together to generalize, but still operate primarily within the realm of semantics.
\begin{table}[ht]
\small
\centering
\begin{tabular}{c c c }
\toprule
&\texttt{SocialChem} &\texttt{MIC}   \\ 
        \midrule
        rot&.389 & .659 \\ 
        moral-rot& .418 & .738 \\ 
        \midrule
        judg&.992 & .770\\ 
        moral-judg& .997 & .835  \\
        \bottomrule
\end{tabular}
\caption{\small The average conditional likelihoods of top-10 generalization-supportive training samples.}
\label{tab:likelihoods}
\end{table}

\begin{figure}[h]
    \centering
    \includegraphics[width=0.75\linewidth]{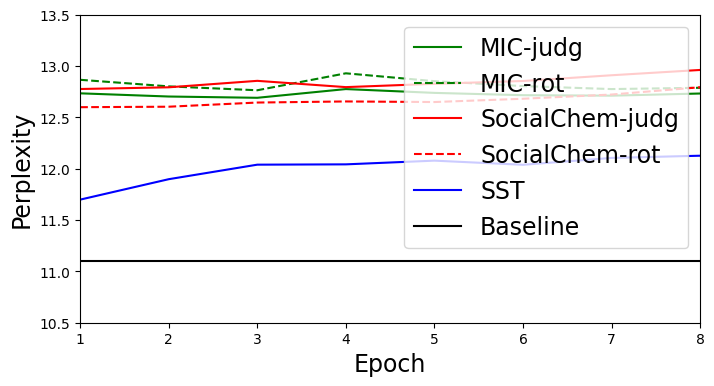}
    \caption{\small Perplexity for Mistral. Baseline indicates the Perplexity of the LLMs without any fine-tuning.}
    \label{fig:perplexisity}
\end{figure}

Recall that, in Section~\ref{sec:preliminary}, we demonstrate that the generalization and convergence behavior of moral foundation classification is different from SST due to the pragmatic delimma. 
Similarly, we also argue that the pragmatic nature of morality would be more negative to the language modeling capability of LLMs than that from SST.
Figure~\ref{fig:perplexisity} presents the perplexity evaluation results, acquired through the OpenWebText datset~\cite{Gokaslan2019OpenWeb}, of Mistral models fine-tuned with different strategies.
It is obvious that morality-relevant tasks introduce more perplexity than SST.

In summary, \textit{while the optimal fine-tuning strategies improve performance on both tasks, this improvement remains within the realm of distributional semantics, and the pragmatic dilemma persists}.

\section{Discussion\label{app:discussion}}
Generalization remains a significant challenge in the acquisition of moral reasoning, and no optimal solution has yet been identified. Recently,~\citet{jiang2025investigating} proposed a hybrid approach that combines bottom-up and top-down methods. However, their method still relies on a substantial number of training samples. 
\citet{bergen2016pragmatic} demonstrated that pragmatic reasoning can be approximated through semantic inferences, highlighting a linguistic foundation for this connection. Nevertheless, how to formally structure a semantic inference framework for moral reasoning remains an open question.
One promising direction is to ground such a framework in the human moral decision-making process. \citet{kumar2025rules} introduced the first benchmark in the NLP community focused on how humans make moral decisions. Their benchmark is based on an intuitionist model: participants are first asked to make a moral judgment and then provide an explanation for their decision. 
This type of annotation presents challenges for LLMs, as human explanations are expressed in free-text form and often lack enough situated semantic information~\cite{sap2022neural}. Despite these difficulties, the benchmark offers a valuable opportunity for exploring methods that aim to derive semantic inferences from human rationales, potentially bridging the gap in pragmatic reasoning for morality.

\citet{sap2019social} propose a social bias inference framework to identify social implications, but their approach still relies heavily on semantics-level signals, as the social implications in their work are primarily associated with explicit toxicity.
\citet{chen2025pragmatic} introduce a pragmatic inference framework targeting implicit toxicity, where metaphorical expressions of toxicity pose particular challenges for LLMs.
They provide empirical evidence that their framework enhances the detection of implicit toxicity in off-the-shelf LLMs, though it still depends on the powerful distributional semantics captured by these models.
We argue that the study of moral reasoning acquisition should focus on smaller LLMs with 3B–8B parameters, as these models are less powerful than very large ones yet still exhibit satisfactory instruction-following capabilities.
The powerful distributional semantics of large models make it difficult to disentangle whether performance gains stem from the proposed pragmatic inference framework or from learned statistical correlations between situations and moral reasoning objectives.

\section{Conclusion}
In this paper, we answered the question \textit{can current learning paradigms enable LLMs to acquire moral reasoning?} 
Based on distributional semantics and the pragmatic nature of morality, we demonstrate that 
(1) the pragmatic dilemma of LLMs make them inefficient in moral reasoning acquisition tasks; 
(2) the improved performance still stems from the realm of distributional semantics;
(3) the current learning paradigm for moral reasoning acquisition impairs LLMs’ language modeling capability more than semantics-level tasks.
We conclude that the pragmatic dilemma is the primary bottleneck for moral reasoning acquisition.
\section*{Limitations}
In this draft, we focus only on moral situations with a single underlying moral foundation. However, in real-world scenarios, moral situations often involve multiple moral foundations, which we leave for future research.
Additionally, while the tasks considered in this paper reflect fundamental aspects of moral reasoning, a deeper analysis of how the pragmatic dilemma manifests in recently proposed social sandbox systems would be a valuable direction for future study.

\bibliography{custom}
\appendix

\section{Appendix\label{sec:appendix}}

\subsection{Additional Related Works\label{sec:relatedworks}}
Machine ethics~\cite{anderson2011machine,tolmeijer2020implementations,nath2020problem,allen2006machine} has been a long-standing research topic for hardware and software systems, with the aim of maximizing their benefits while minimizing societal risks.
Recently, we have witnessed the progress of Artificial Intelligence (AI), particularly that associated with Large Language Models (LLMs), changing the world.
Ensuring LLMs will acquire an understanding of ethics to prevent them from making harmful decisions has become a serious research problem for both academia and industry.
Dating back to the 1940s, the Three Laws of Robotics~\cite{asimov1941three} were proposed to ensure that robots do not cause harm to humans.
Since then, machine ethics has been explored by researchers in philosophy, psychology, and cognitive science. However, it remains a significant challenge for AI, as even coherent and diverse language generation poses difficulties.
The widespread deployment of LLMs opens the door for AI researchers to pursue ethics acquisition due to their strong semantic modeling capability. 

Numerous studies have attempted to evaluate the moral and ethical orientations encoded in LLMs through empirical experiments. \citet{bonagiri2024sage} demonstrates that model performance and moral consistency are independent of one another, while \citet{abdulhai2023moral} investigates whether LLMs exhibit biases toward specific moral principles. \citet{scherrer2024evaluating} proposes a statistical method to assess the moral values encoded in LLMs, and \citet{zhang2023measuring} introduces a metric to determine whether LLMs understand ethical values both in terms of “knowing what” and “knowing why.” Collectively, these studies highlight that LLMs lack consistent moral or ethical orientations across different scenarios.
Enabling LLMs to acquire ethical values is a formidable challenge, not only because ethical AI operates at the level of pragmatics~\cite{awad2022computational}, but also due to the philosophical complexities surrounding the proper representation of human ethics~\cite{zhixuan2024preferencesaialignment}. Progress has been made, albeit only partially. 

\subsection{Hyperparameters for the Bert Classifier\label{appendix:hyperparams4clf}}
Hyperparameters are available in Table~\ref{tab:optimhyperparam}.
\begin{table}[h]
    \centering
    \begin{tabular}{cc}
    \toprule
     \textbf{Hyperparameters} & \textbf{Setting} \\
     \midrule
     \textbf{Optimizer} & \text{AdamW}\\
     \textbf{Adam $\beta_1$} & \text{0.9} \\
     \textbf{Adam $\beta_2$} & \text{0.98} \\
     \textbf{Adam $\epsilon$} & \text{1e-3} \\
     \textbf{Learning rate for BERT} & \text{5e-5} \\
     \textbf{Learning rate for classifier layer} & \text{1e-2} \\
     \textbf{Maximum training epochs} & \text{10} \\
     \textbf{Weight decay} & \text{0.01} \\
     \textbf{Batch size} & \text{32}\\
     \textbf{Seed} & 1,2,3,4,5\\
     \bottomrule
    \end{tabular}
    \caption{Hyperparameter Settings for the AdamW Optimizer.}
    \label{tab:optimhyperparam}
\end{table}
\subsection{Re-categorization of Moral Foundation Labels\label{appendix:redist_moral}}
For \textbf{MIC}, we label samples with the moral foundation of \textit{Care} as 0, and those with the foundations of \textit{Fairness, Liberty, Authority}, and \textit{Loyalty} as 1.
For \textbf{SocialChem}, samples classified under \textit{Loyalty-Betrayal} are labeled as 0, while those falling under \textit{Fairness-Cheating}, \textit{Care-Harm}, \textit{Sanctity-Degradation}, and \textit{Authority-Subversion} are labeled as 1.
\subsection{Experimental Settings for Fine-tuning\label{appendix:ethicaltuning}}
\begin{center}
\begin{tcolorbox}[colback=black!2,colframe=black,width=7.6cm,arc=1mm,boxrule=0.5pt]
\small
Prompting format moral-rot-judgment \\ \\ Situation: \{\#SITUATION\}\\
Moral Foundation: \{\#MORAL\_FOUNDATION\}\\
Rule of Thumb: \{\#RoT\}\\
Ethical Judgment: \{\#judgment\}\\

LoRA hyperparameters\\
rank: 64\\
lora alpha: 16\\
lora dropout: 0.1\\
target modules: q\_proj, k\_proj, v\_proj, o\_proj\\
batch size: 16\\
learning rate: 5e-5
\end{tcolorbox}
\end{center}

\subsection{Mechanistic Analysis to Fine-tuned Llama3\label{appendix:mech_llama3}}
\begin{figure*}
    \centering
    \subfloat[\tiny RoT Generation for Socialchem]{\includegraphics[width=0.5\linewidth]{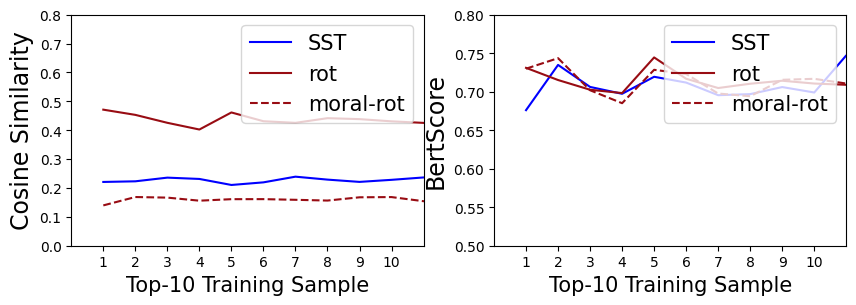}}
    \hfill
    \subfloat[\tiny Ethical Judgment Prediction for SocialChem]{\includegraphics[width=0.5\linewidth]{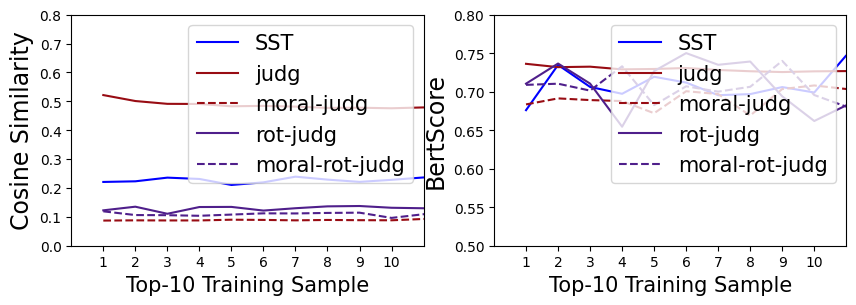}}
    \hfill
    \subfloat[\tiny RoT Generation for MIC]{\includegraphics[width=0.5\linewidth]{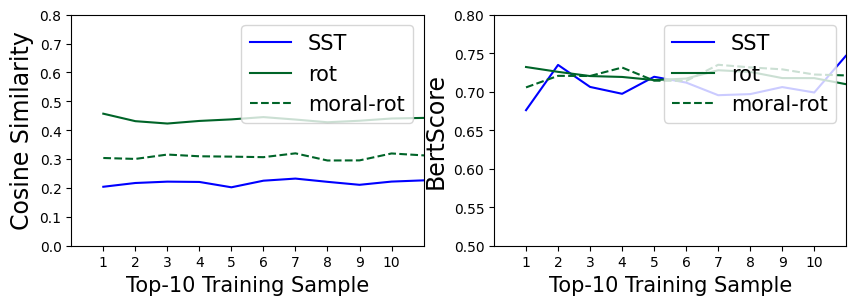}}
    \hfill
    \subfloat[\tiny Ethical Judgment Prediction for MIC]{\includegraphics[width=0.5\linewidth]{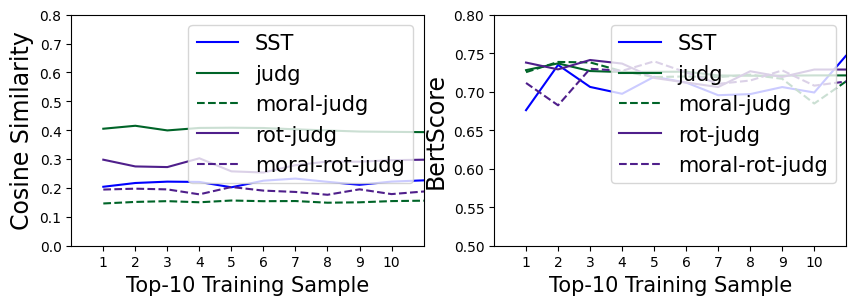}}
    \hfill
    \caption{\small Top-10 Generalization-Supportive Training Samples Analysis for Fine-tuned Llama3 Through the Introduced Fine-tuning Strategies.}
    \label{fig:mech_llama3}
\end{figure*}

\begin{figure*}[h]
    \centering
    \begin{minipage}{0.24\textwidth}
        \centering
        \includegraphics[width=\linewidth]{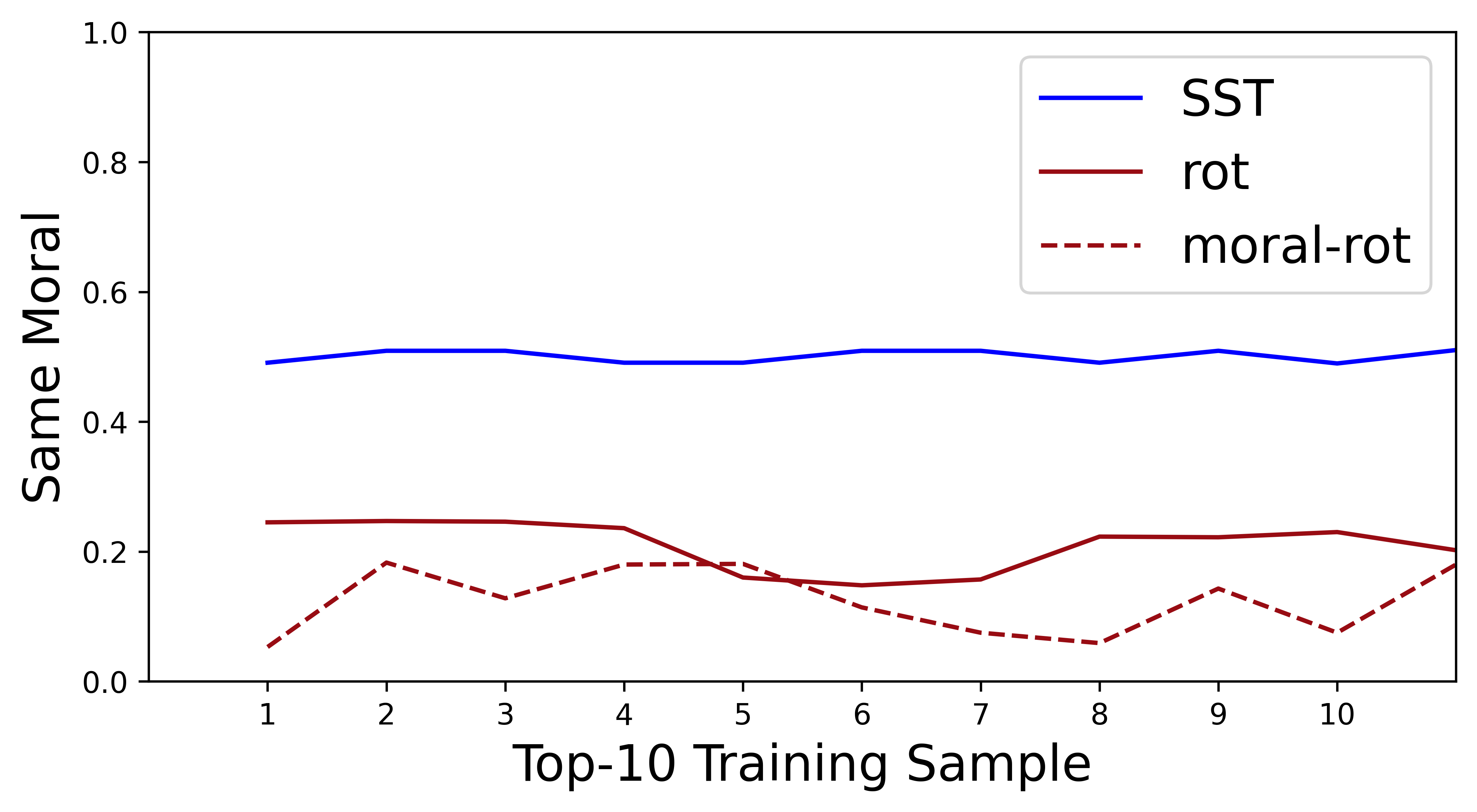}
        \caption{\small RoT in SocialChem}
    \end{minipage}
    \hfill
    \begin{minipage}{0.24\textwidth}
        \centering
        \includegraphics[width=\linewidth]{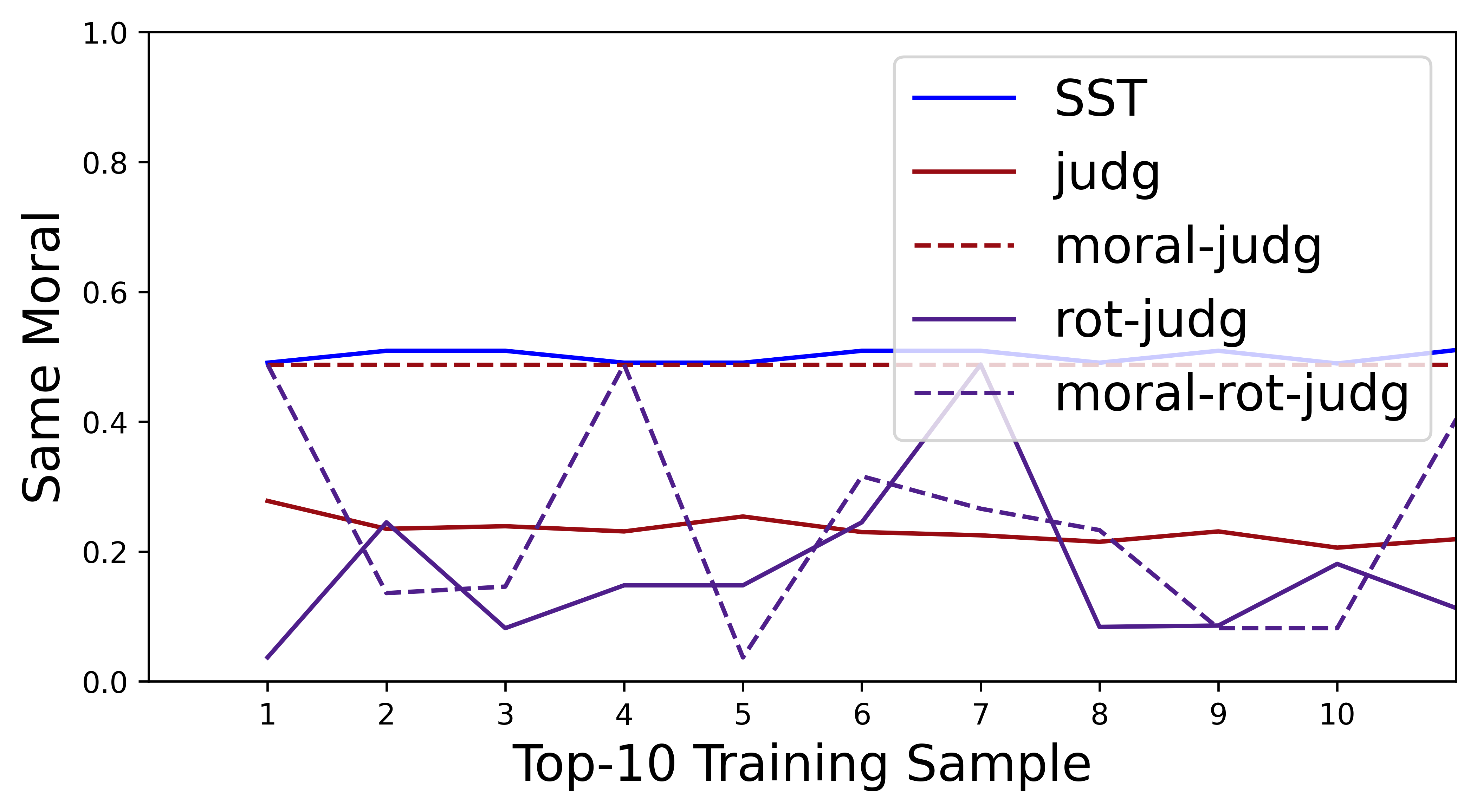}
        \caption{\small Ethical Judgment Prediction in SocialChem}
    \end{minipage}
    \hfill
    \begin{minipage}{0.24\textwidth}
        \centering
        \includegraphics[width=\linewidth]{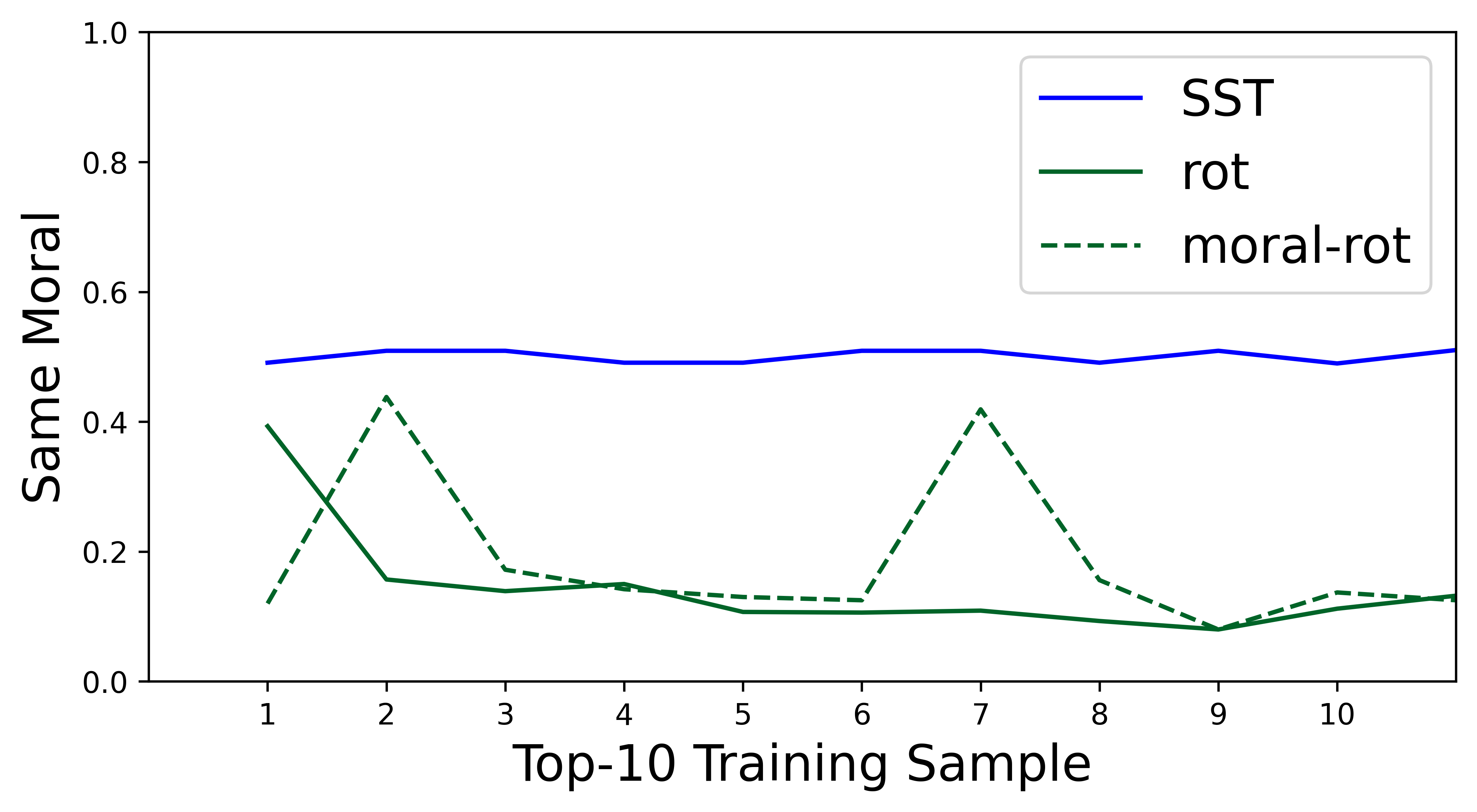}
        \caption{\small RoT in MIC}
    \end{minipage}
    \hfill
    \begin{minipage}{0.24\textwidth}
        \centering
        \includegraphics[width=\linewidth]{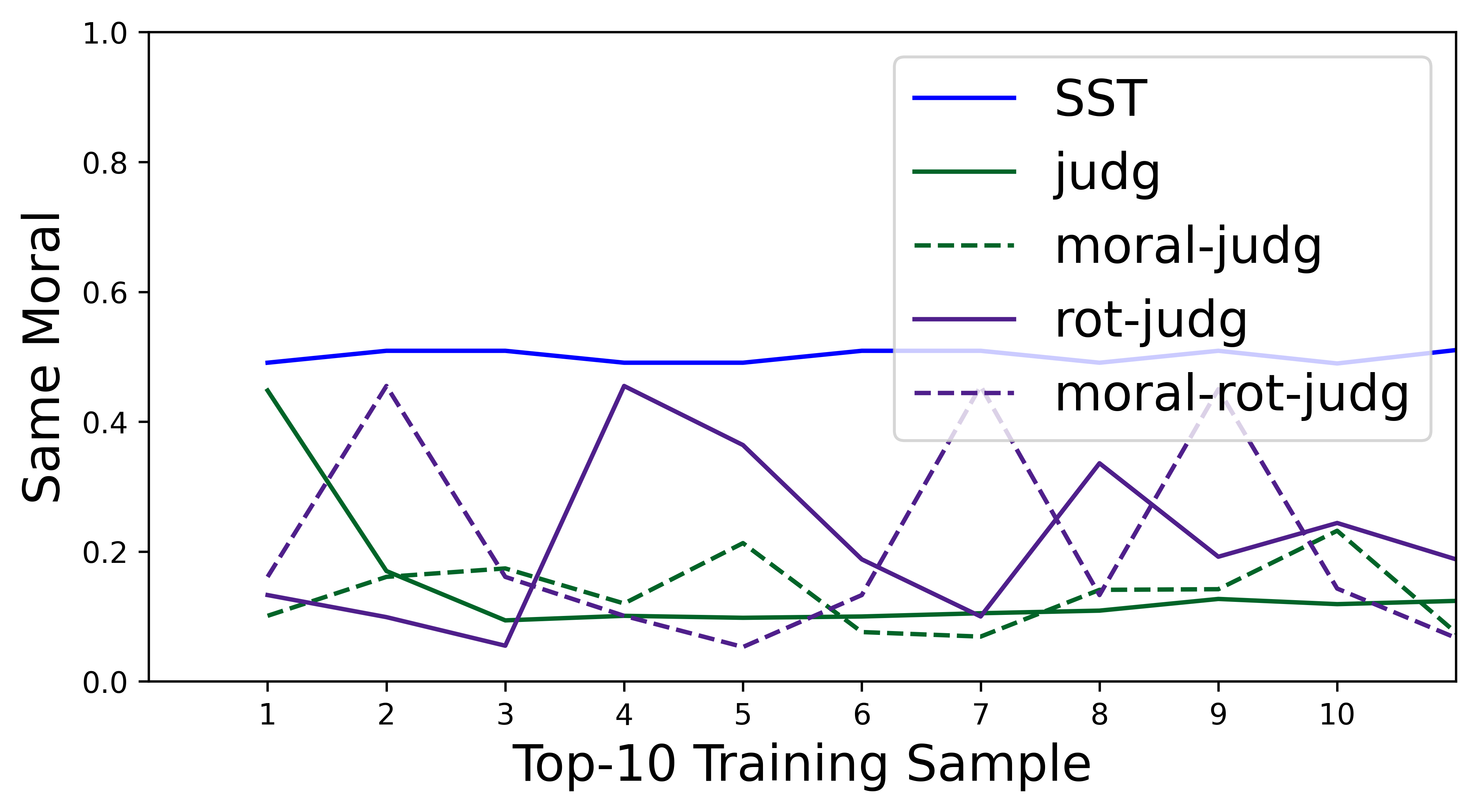}
        \caption{\small Ethical Judgment Prediction in MIC}
    \end{minipage}
    \caption{\small Same Moral Ratio for Fine-tuned Llama3.}
    \label{fig:same_moral_llama3}
\end{figure*}

Table~\ref{fig:mech_llama3} introduces the fine-tuning results for the Llama3 model. 
Different from Mistral, introducing additional information of the moral foundations and RoT do not always contribute to better performance.
For the SocialChem benchmark, the baseline fine-tuning strategy outperforms other strategies, albeit by a very narrow margin.
This aligns with the generalization mechanism illustrated in Figure~\ref{fig:mech_llama3}. Unlike Mistral, the introduction of moral foundations and RoT does not reduce cosine similarity or BertScore.
Figure~\ref{fig:same_moral_llama3} shows the ratio of the same moral foundation among top 10 generalization-supportive training moral situations, and the behavior of Llama3 is the same as Mistral.
In summary, the pragmatic dilemma still persists for the Llama3 model and is even worse than that of the Mistral model.
\begin{table*}[h]
\small
\centering
\begin{tabular}{c c c c c| c c c c c}
\toprule
\texttt{SocialChem} &BertScore & Rouge1 & Rouge2 & RougeL & \texttt{MIC} &BertScore & Rouge1 & Rouge2 & RougeL \\ 
        \midrule
        \textbf{rot}&.8222 & .358 & .151 & .343& rot&.814 & .365 & .152 & .332 \\ 
        moral-rot&.8217 & .356 & .152 & .340 & \textbf{moral-rot}&.818 & .365 & .168 & .352 \\ 
        \midrule
        \textbf{judg}&.759 & .440 & .313 & .440 & judg &.684 & .109 & .000 & .109 \\ 
        \underline{moral-judg} & .757 & .411 & .285 & .411 & \underline{moral-judg} &.751 & .254 & .000 & .254 \\
        rot-judg & .755 & .400 & .264 & .400 & rot-judg &.660 & .061 & .000 & .061 \\
        \makecell{moral-rot\\-judg}&.752 & .370 & .248 & .370& \textbf{\makecell{moral-rot\\-judg}}&.762 & .314 & .000 & .314 \\
        \bottomrule
\end{tabular}
\caption{\small Performance of Fine-tuned \texttt{Llama3} Model Across Various Fine-tuning Strategies for Each Benchmark. The best fine-tuning strategy is highlighted in \textbf{bold} and the second best strategy is \underline{underlined}. For MIC, incorporating additional information, such as moral foundations, during fine-tuning enhances performance; however, this effect is not observed for SocialChem.}
\label{tab:ethicaltuning4llama3}
\end{table*}

\begin{table*}[t]
\small
\centering  
\begin{tabular}{|l|l|}
\toprule
Moral Foundation Branches & Brief Description \\
\midrule
\makecell{Care\\Harm} & \makecell{Demonstrates care, generosity, compassion, and empathy, \\while showing sensitivity to others’ suffering and upholding the principle of avoiding harm.}\\ 
\midrule
\makecell{Fairness\\Cheating} & \makecell{Encompasses fairness, justice, reciprocity, altruism, rights, \\autonomy, equality, proportionality, and the rejection of cheating.} \\
\midrule
\makecell{Loyalty\\Betrayal} & \makecell{Emphasizes group affiliation, solidarity, patriotism, \\and self-sacrifice, while prohibiting betrayal.} \\
\midrule
\makecell{Authority\\Subversion}& \makecell{Upholding social roles, respecting authority and \\traditions, valuing leadership, and prohibiting rebellion.} \\
\midrule
\makecell{Purity (Sanctity)\\Degradation} & \makecell{Reverence for the sacred, purity, religious principles guiding life, \\and prohibitions against violating the sacred.} \\
\bottomrule
\end{tabular}

\caption{Brief Descriptions of the Moral Foundations. Each foundation has two aspects representing positive and negative perspectives of that moral foundation branch. Please refer to~\citet{atari2023morality} for the most up-to-date list of moral foundations and their descriptions.} 
\label{appendix:MFs}
\end{table*}

\subsection{Top-50\label{app:top100}}
In Figure~\ref{fig:same_moral_mistral}, we show only 10 generalization-supportive samples. Here, we demonstrate that the characteristics of all top-50 generalization-supportive training samples are closely aligned with those of the top 10 reported in that figure.

Mistral-SocialChem-RoT:[0.138, 0.16, 0.221, 0.193, 0.189, 0.185, 0.21, 0.193, 0.17, 0.178, 0.18, 0.176, 0.181, 0.191, 0.187, 0.157, 0.161, 0.145, 0.163, 0.133, 0.15, 0.181, 0.152, 0.162, 0.18, 0.163, 0.173, 0.16, 0.158, 0.186, 0.176, 0.178, 0.17, 0.185, 0.171, 0.169, 0.165, 0.194, 0.191, 0.173, 0.19, 0.173, 0.188, 0.192, 0.188, 0.195, 0.189, 0.19, 0.195, 0.17] with mean value of 0.17636

Mistral-Socialchem-MoralRoT: [0.051, 0.243, 0.238, 0.214, 0.079, 0.245, 0.244, 0.244, 0.241, 0.216, 0.072, 0.133, 0.204, 0.276, 0.137, 0.179, 0.178, 0.115, 0.049, 0.151, 0.152, 0.16, 0.089, 0.048, 0.186, 0.141, 0.126, 0.137, 0.146, 0.047, 0.045, 0.041, 0.122, 0.156, 0.143, 0.084, 0.237, 0.232, 0.135, 0.099, 0.09, 0.207, 0.371, 0.169, 0.23, 0.127, 0.093, 0.199, 0.164, 0.163] with mean of 0.15696


\end{document}